\documentclass[letterpaper]{article} 
\usepackage{aaai24}  
\usepackage{times}  
\usepackage{helvet}  
\usepackage{courier}  
\usepackage[hyphens]{url}  
\usepackage{graphicx} 
\usepackage{natbib}  
\usepackage{caption} 
\usepackage{amsmath,amsfonts}
\usepackage{array}
\usepackage{multirow}
\usepackage{booktabs} 
\usepackage{url}
\usepackage{color}

\usepackage{algorithm}
\usepackage{algorithmic}

\usepackage{newfloat}
\usepackage{listings}
\DeclareCaptionStyle{ruled}{labelfont=normalfont,labelsep=colon,strut=off} 
\floatstyle{ruled}
\newfloat{listing}{tb}{lst}{}
\floatname{listing}{Listing}
\title{Direct May Not Be the Best: An Incremental Evolution View of Pose Generation}
\author{
    Yuelong Li\textsuperscript{\rm 1,2},
    Tengfei Xiao\textsuperscript{\rm 3},
    Lei Geng\textsuperscript{\rm 4},
    Jianming Wang\textsuperscript{\rm 2}\thanks{Corresponding author.}\\
}
\affiliations{
    \textsuperscript{\rm 1}School of Artificial Intelligence, Tiangong University, Tianjin, 300387, China\\
    \textsuperscript{\rm 2}Tianjin Key Laboratory of Autonomous Intelligence Technology and Systems, Tiangong University, Tianjin, 300387, China\\
    \textsuperscript{\rm 3}School of Software, Tiangong University, Tianjin, 300387, China\\
    \textsuperscript{\rm 4}School of Life Sciences, Tiangong University, Tianjin, 300387, China
    
    liyuelong@pku.edu.cn, newtnt121@qq.com, \{genglei,wangjianming\}@tiangong.edu.cn
}

\usepackage{bibentry}

\definecolor{HighLight}{RGB}{0,31,141}
\begin{document}

\maketitle


\begin{abstract}
Pose diversity is an inherent representative characteristic of 2D images. Due to the 3D to 2D projection mechanism, there is evident content discrepancy among distinct pose images. This is the main obstacle bothering pose transformation related researches. To deal with this challenge, we propose a fine-grained incremental evolution centered pose generation framework, rather than traditional direct one-to-one in a rush. Since proposed approach actually bypasses the theoretical difficulty of directly modeling dramatic non-linear variation, the incurred content distortion and blurring could be effectively constrained, at the same time the various individual pose details, especially clothes texture, could be precisely maintained. In order to systematically guide the evolution course, both global and incremental evolution constraints are elaborately designed and merged into the overall framework. And a novel triple-path knowledge fusion structure is worked out to take full advantage of all available valuable knowledge to conduct high-quality pose synthesis. In addition, our framework could generate a series of valuable by-products, namely the various intermediate poses. Extensive experiments have been conducted to verify the effectiveness of the proposed approach. Code is available at \url{https://github.com/Xiaofei-CN/Incremental-Evolution-Pose-Generation}.    
\end{abstract}


\section{Introduction}

At present, 2D image is still the most widely used visual information transmission and storing carrier, which is structure simple, display intuitive, and manufacturing efficient. But since 2D images are only projections of genuine 3D objects to 2D planes, pose variation is an intrinsic characteristic of this data category. Clearly, the dimension degrading mechanism and object self-occlusion imply the projection process is irreversible, and thus adjust or normalize object pose becomes a quite tough mission that bothers tremendous 2D image based compute vision tasks, such as object detection, tracking, recognition, and understanding. Among massive existing objects, with no doubt, human body is one of the most challenging one, due to prominent shape flexibility, diverse subtle clothes texture, and plenty of potential pose categories. In this paper, we specifically focus on human pose transformation and generation. 







Clearly, human pose variation involves dramatic non-linear visual content variation modeling, which is still theoretically difficult by now. Thus, in the past, unlike relatively rigid human face \cite{MLS2012,3Dmodel,Dictionary}, pose generation is always a much tougher task and effective solutions are relatively rare. Real breakthrough developments come until the era of deep neural networks, which is an excellent modeling technique with overwhelming advantages over traditional modeling approaches, as to description capacity, robustness, flexibility, and especially complex non-linear modeling. 






\begin{figure}[t]
\centering
\includegraphics[width=\linewidth]{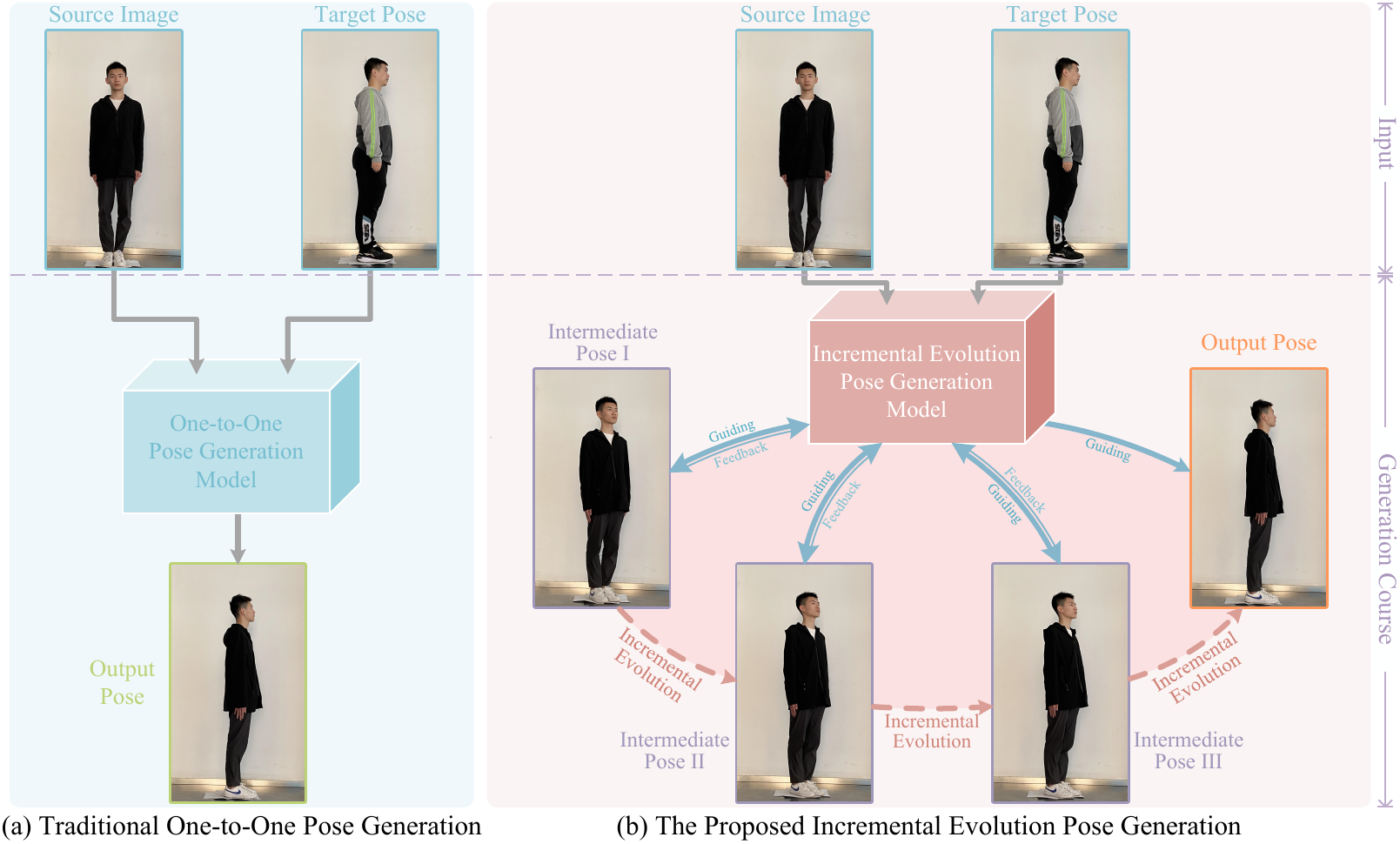}
\vspace{-5mm}
\caption{The basic flows of classical one-to-one pose generation (a) and proposed incremental evolution synthesis (b).}
\label{fig_gradual_process_illustration}
\vspace{-5mm}
\end{figure}


In the past few decades, accompanied by the booming of deep learning techniques, a number of pose synthesizing approaches have been worked out. Competitive process originated Generative Adversarial Networks (GANs) \cite{GAN} is a powerful modeling architecture famous for prominent missing information compensation and complex content generation capability. Thus, this structure is an evident characteristic of numerous pose generation approaches \cite{BiGraphGAN,PAIG,AD-GAN,MSAGPT,PATN,XingGAN,DPTN,PISE}. 
Knowledge transferring in feature space is also a popular way to realize pose transformation \cite{PAIG,SPGNet,PINet}, where multi-stage information fusing is widely used to release the stress of missing information modeling. In additional, newly developed attention centered Transformer structure has also proved its effectiveness for robust modeling human pose variation \cite{CASD,DPTN}.  


However, despite these promising advances, generally, robust pose generation is still a tough mission, and we could find from these reported papers that always perfect synthesizing performance are hardly achieved, especially when the pose distinction is huge. As to the fundamental reason, in our opinion, that's because the dramatic non-linear variance modeling of visual content is still theoretically challenging for this moment. Thus, how to effectively conduct pose transforming modeling and achieve high-quality synthesizing are our core objectives in this paper. 



As to the huge non-linear content distinctions among various human poses, we don't think traditional one-to-one straightforward transfer modeling is the unique wise answer. Bypassing the theoretical intractability and finding out an indirect solution may also be a good way to the final goal. In this paper, we designed a slight pose transformation unit centered gentle incremental evolution synthesis framework, where no dramatic pose difference has to be directly handled, and hence, the fundamental modeling difficulty hovering pose transforming mission has been indirectly "solved" to some extent. 
The core idea is intuitively demonstrated in Figure \ref{fig_gradual_process_illustration}. Compared with traditional rush one-to-one generation, our entire generation framework is comprised by a series of tightly related incremental evolution synthesizing units which are rigorously controlled by global guidance and incremental feedback enhancing. We will specifically introduce the technical details in the method section.




The main contributions of this paper are as follows:
\begin{itemize}
  \item A gentle incremental evolution angle of view is proposed to understand dramatic variance pose generation. 
  \item Two angles evolution course guidance and a novel triple-path knowledge fusion structure are worked out and integrated to boost entire pose evolution synthesizing flow. 
  \item We provide an indirect solution to relieve the theoretical challenge of direct modeling wide non-linear discrepancy.
  \item Besides the main objective, proposed pose synthesizing approach could also generate a series of valuable intermediate poses as by-products, which may be beneficial to plenty of related tasks in an era of data being king.
\end{itemize}     




\section{Related Works}
In the past decades, the intrinsic flexibility of various human body attracts the attention of a great many of researchers within computer vision society. But due to the inherent challenging of complex non-linear modeling, 
it was not until the era of deep learning that genuine breakthroughs were made.

As one of the most prominent visual content generation framework, adversarial mechanism originated GANs \cite{GAN} is widely adopted to generate various body poses. XingGAN \cite{XingGAN} introduces a crossing structure where both shape and appearance information are extensively fused, and both shape-guided and appearance-guided discriminator are used to lead the adversarial procedure. To conduct semantics guided pose generation, Li \textit{et al.} \cite{SPMPG} proposed a two paths encoding-decoding framework where both image and pose path are mixed through multi-stage attention. Besides, they also worked out a multi-scale discriminator. 
Roy \textit{et al.} \cite{MSAGPT} introduced attention links at every resolution level of the encoder and decoder. The discriminator takes two channel-wise concatenated images as input and conducts patch based adversarial competition. 
BiGraphGAN \cite{BiGraphGAN} captures the crossing long-range relations between source and target pose through bipartite to mitigate the challenges caused by pose deformation. Khatun \textit{et al.} \cite{PAIG} put forward a network structure to transfer subject pose through attention. Here, both an appearance discriminator and a pose discriminator are used to guide the synthesizing training. Based on parsing map, Zhang \textit{et al.} \cite{PISE} proposed a joint global and local per-region encoding and normalization mechanism to compensate invisible regions. 

The core destination of pose generation is to faithfully transfer source visual content into target pose, and hence how to realize information migration and fusion are critical. Zhang \textit{et al.} \cite{PINet} designed a two levels hierarchical synthesizing mechanism, where the first level is mainly in charge of transferring target pose into source semantics, while the second level engages to the further merging of image level knowledge. With similar overall two-stages structure, Lv \textit{et al.} \cite{SPGNet} proposed a novel information fusing strategy, namely region-adaptive normalization, where per-region styles are used to guide the target appearance generation.   


\begin{figure*}[t]
\centering
\includegraphics[width=\linewidth]{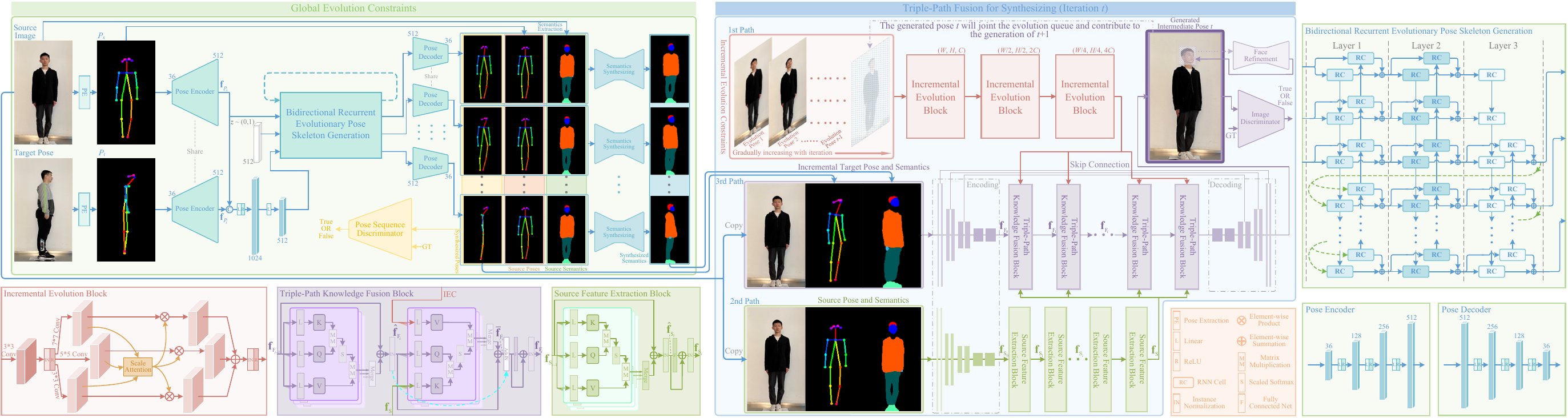}
\vspace{-5mm}
\caption{Overview of the proposed framework, where top-left is the dual input, namely the source image and target pose. The upper left of the figure demonstrates the recurrently progressive generation of global evolution constraints. The middle part shows the triple-path knowledge fusion based pose synthesizing, at iteration \textit{t}, which is the core unit structure of proposed incremental evolution pose generation.
}
\label{fig_overallflow}
\vspace{-3mm}
\end{figure*}

Recently, dense attention centered Transformer structure is enrolled to synthesize various poses as well. Zhou \textit{et al.} \cite{CASD} introduced a cross attention based style distribution block which could effectively fuse the source semantic styles with the target pose. Zhang \textit{et al.} \cite{DPTN} proposed a siamese structure composed of source-to-source self-reconstruction and source-to-target generation, where a Transformer module is worked out to mix the information coming from dual branches. Bhunia \textit{et al.} \cite{PIDM} put forward a texture diffusion module based on cross attention to model the correspondences between appearance and pose information available in source and target images. They demonstrated that the denoising diffusion models can be applied to pose image synthesis.


\section{Our Method}
\label{sec:method}



Facing the intrinsic challenge of huge non-linear deformation representation, rather than direct transforming modeling, we explored a novel incremental evolution based gentle solution which is characterized by a series of rigorously constrained slight pose evolution units. The overall  framework is shown in Figure \ref{fig_overallflow}. Here two angles constraints (global and incremental) are introduced to strictly guide and regularize the evolution course, and an integrated triple-path knowledge fusion structure is designed to achieve incremental high-quality pose synthesizing. 



\subsection{Global Evolution Constraints}   
Though changing classical one-to-one generation manner into gentle evolution framework could effectively relieve the theoretical challenge of direct modeling huge pose variance, enrolling multiple sequential components may incur extra evolution instability. 
Thus, we impose rigorous global constraints to ensure the overall incremental evolution trajectory functionally work well towards the ultimate target. It is carried out through incremental pose skeleton and image semantics landmarks. It is well known that skeleton and semantics are the most commonly pose description tools for visual content transferring \cite{BiGraphGAN,PAIG,AD-GAN,MSAGPT,PATN,XingGAN,DPTN,PISE,SPGNet,PINet,CASD,SPMPG,PIDM}, and hence the overall synthesizing course could be tightly controlled through building global guiding skeleton and semantics evolution sequences. 
The detailed process is demonstrated in the upper left of Figure \ref{fig_overallflow}.



The pose skeletons  $P_s$ and $P_t$ are first extracted from corresponding input images, which is conducted mainly based on OpenPose \cite{OpenPose}. Then we construct a three layers fully connected network as Pose Encoder to acquire their  512 dimensions features $\textbf{f}_{P_s}$ and $\textbf{f}_{P_t}$. Considering the incrementally evolutionary nature of proposed synthesizing framework, we design a recurrent neural network (RNN) originated structure. In detail, to accurately model the gradual evolution course, we adopt bidirectional relationships and cascade three layers. This multiple layers structure contributes to the comprehensive modeling of the evolution rules. The output of the $l$th layer at time step $t$ is, 
\begin{equation}
O^l_t = \Phi_{W_f}(X^l_t, \overrightarrow{H}^l_{t-1})\oplus \Phi_{W_b}(X^l_t, \overleftarrow{H}^l_{t+1}),
\end{equation}
 where $\Phi_{W_f}$ and $\Phi_{W_b}$ denote trainable forward and backward directional RNN cell functions,  $X^l_t$ represents the layer input, $\overrightarrow{H}^l_{t-1}$ and $\overleftarrow{H}^l_{t+1}$ are the former hidden states of both directions respectively. The recurrent structure accepts source feature $\textbf{f}_{P_s}$, random vector $\textbf{z}\sim(0,1)$, and the mixing of $\textbf{f}_{P_s}$ and  $\textbf{f}_{P_t}$ through two linear layers, as the starter. The recurrent outputs are further processed by a three layers linear network (Pose Decoder) to obtain the desired global guiding pose skeletons. Moreover, the generation flow is enhanced by a adversarial pose sequence  discriminator to guarantee the overall quality.    

After obtaining pose constraints, the semantics constraints are generated based on the parsing generator proposed in \cite{PISE}, where source pose and semantics, and desired pose are the combined input. Here the source semantics is acquired based on \cite{SemanticExtraction}.    

\subsection{Incremental Evolution Constraints}
Besides the mentioned global constraints, we want to further improve the evolution stability by physically modeling the incremental tendency, and thus an explicit incrementally progressive guiding branch is worked out, as shown in the middle upper of Figure \ref{fig_overallflow} (pink color). In this branch, all intermediates generated before current iteration $t$ are sequentially integrated as the input, so that previous evolution tendency could be directly collected and used to guide current generation. Here, the output of current iteration will join next iteration's generation to update and enrich the evolution tendency queue, as instantaneous feedback. 

The tendency learning is mainly conducted by three stacked Incremental Evolution (IE) Blocks, which are designed to effectively extract multi-level visual knowledge and adaptively model evolution patterns. 
In detail, each IE block is constructed based on a bunch of full-channel convolution units and scale-wise attention fusion under splitting analysis and then adaptively reassembling overall topology. The multiple size kernels (3*3, 5*5, and 7*7) are integrated through learnable scale attention to systematically acquire various level visual features, while full-channel knowledge analysis is responsible for evolution tendency exploration and extraction. The stacking sequentially evolves the feature matrix from $(W,H,C)$, $(W/2,H/2,2C)$, to $(W/4,H/4,4C)$. Finally, the obtained evolution tendency will be densely mixed into the main synthesizing flow to fully constrain the incremental evolution course, as Incremental Evolution Constraints. 




\subsection{Triple-Path Knowledge Fusion for Synthesizing}
In order to comprehensively integrate all obtained valuable information to realize high-quality pose generation, an elaborately designed triple-path knowledge fusion mechanism is worked out. Here, the incremental evolution constraints (1st Path, the middle upper of Figure \ref{fig_overallflow}), and source image, pose and semantics (2nd Path, the middle lower of Figure \ref{fig_overallflow}), act as all-round penetrators to the entire synthesizing flow, while source image combined with the global constraints, namely current incremental target pose and semantics (3rd Path, the center of Figure \ref{fig_overallflow}), play a role as path start input. The middle part of Figure \ref{fig_overallflow} detailed demonstrates this cascaded multi-source fusion structure at evolution iteration \textit{t}. Inspired by Transformer \cite{Attention}, the fusion process is designed based on attention exploration as the basic element. Here, both the source and incremental target information are elementarily processed by classical encoding-decoding structure \cite{DPTN}. 

After encoding, the acquired source features $\textbf{f}_{\rm{S}_0}$ are further processed through a series of Source Feature Extraction (SFE) blocks, as shown in the middle bottom of Figure \ref{fig_overallflow}. In each SFE, the input features are first linearly projected into three embedding space, namely, Key, Query, and Value, and then their intrinsic attention relationships are extensively explored. Multi-head mechanism is enrolled also to further boost information diversity. Specifically, the acquisition after SFE block $i$ is, 
\begin{align}
\textbf{f}_{\rm{S}_\textit{i}} &=IN\left[FCN(\hat{\textbf{f}}_{\rm{S}_\textit{i}})\oplus\hat{\textbf{f}}_{\rm{S}_\textit{i}}\right],\\
\hat{\textbf{f}}_{\rm{S}_\textit{i}} &=IN\left[\textbf{f}_{\rm{S}_{\textit{i}-1}}\oplus \mathop{Merge}\limits_{j=1:n_{MH}}\left(Attn\left(L^j_{K/Q/V}(\textbf{f}_{\rm{S}_{\textit{i}-1}})\right)\right)\right],
\label{eq:attention}
\end{align}
where \textit{IN} denotes instance normalization, \textit{FCN} is fully connected network, $n_{MH}$ represents multi-head number, $Attn$ indicates attention computing, and $L_{*}$ is linear projection. $\textbf{f}_{\rm{S}}$ is the final output of this path. 


The triple-path fusion task is mainly conducted by cascaded Triple-Path Knowledge Fusion (TPKF) blocks, as shown in the center of Figure \ref{fig_overallflow}. In each TPKF block, the input is extensively explored through overall attention the same as that conducted through Eq (\ref{eq:attention}) to obtain $\hat{\textbf{f}}_{\rm{F}_\textit{i}}$. Then incremental evolution constraints (IEC), $\hat{\textbf{f}}_{\rm{F}_\textit{i}}$, and $\textbf{f}_{\rm{S}}$, are respectively projected into Value, Query, and Key space, and their cross-attentional relationships are explored to realize triple-path fusion, 
\begin{equation}
\bar{\textbf{f}}_{\rm{F}_\textit{i}}=\mathop{Merge}\limits_{j=1:n_{MH}}\left[Attn\left(L^j_V({\rm{IEC}}), L^j_Q(\hat{\textbf{f}}_{\rm{F}_\textit{i}}), L^j_K(\textbf{f}_{\rm{S}})\right)\right]\oplus \hat{\textbf{f}}_{\rm{F}_\textit{i}}.
\end{equation}
Here, since all knowledge of previous blocks are transmitted through the main fused feature, $\hat{\textbf{f}}_{\rm{F}_\textit{i}}$, we introduce an extra direct AdaIN \cite{adain} connection to strengthen its function in the fusion flow, namely,
\begin{equation}
AdaIN(\bar{\textbf{f}}_{\rm{F}_\textit{i}}, \hat{\textbf{f}}_{\rm{F}_\textit{i}}),
\end{equation}   
which is a content normalization based on $\hat{\textbf{f}}_{\rm{F}_\textit{i}}$.

The output of this cascaded structure is then decoded to acquired the synthesized pose of this iteration. To further boost the synthesizing performance, the incremental pose generation course would adversarially compete with a single image discriminator. In addition, based on the approach introduced in  \cite{FaceEnhance}, a detail refinement structure is enrolled to improve face quality. 


\subsection{Learning Objectives}
Since the building of global evolution constraints is relatively independent with the main synthesizing flow, we would train both portions separately for the sake of computing efficiency.   

\vspace{1mm}
\noindent{\textbf{Global Evolution Constraints:}} The training course is driven by the combination of skeleton sequence adversarial loss, neighboring consistency, and single pose quality, 
 \begin{equation}
 \mathcal{L}_{GEC} = \lambda_{sadv}\mathcal{L}_{sadv}+\lambda_{ncons}\mathcal{L}_{ncons}+\lambda_{pose}\mathcal{L}_{pose},
 \end{equation}
where $\lambda_{sadv}$, $\lambda_{ncons}$, and $\lambda_{pose}$ are the relative weights.
In detail, $\mathcal{L}_{sadv}$ corresponds to the adversarial competition of global pose evolution sequence,   
\begin{equation}
\mathcal{L}_{sadv} = \mathbb{E}[\log(1-D_S(\tilde{S}_{pose}))]+\mathbb{E}[\log D_S({S}_{pose})],
\end{equation}
where $D_S$ denotes sequence discriminator, $\tilde{S}_{pose}$ and $S_{pose}$ represent the synthesized and ground truth (GT) pose evolution sequence. $\mathcal{L}_{ncons}$ is in charge of the consistency of local neighboring poses, 
\begin{equation}
\mathcal{L}_{ncons} =  \mathbb{E}\|\tilde{P}_n-\tilde{P}_{n+1}\|_2^2,
\end{equation}
where $\tilde{P}$ represents generated pose. In addition, $\mathcal{L}_{pose}$ guarantees the similarity of each synthesized pose $\tilde{P}$ with corresponding GT pose $P$,
\begin{equation}
\mathcal{L}_{pose} = \mathbb{E} \|P-\tilde{P}\|_2^2.
\end{equation} 


\noindent{\textbf{Pose Image Synthesizing: }} We design to train the whole process with a uniform integrated synthesizing objective, 
\begin{equation}
\begin{split}
 \mathcal{L}_{PIS}  &= \mathcal{L}_{es}+\mathcal{L}_{sr},\\
\mathcal{L}_{es}    &= \sum_t \big{(}\lambda_{siadv}\mathcal{L}^t_{siadv}+\lambda_{style}\mathcal{L}^t_{style}+\lambda_{per}\mathcal{L}^t_{per}\\
                                 &+\lambda_{img}\mathcal{L}^t_{img}\big{)},
\end{split}
\end{equation}
where under  incremental iteration $t$, $\mathcal{L}^t_{siadv}$ is single image adversarial loss, $\mathcal{L}^t_{style}$ and $\mathcal{L}^t_{per}$ respectively are the style loss \cite{PISE} and perceptual loss \cite{XingGAN}, and $\mathcal{L}^t_{img}$ measures the $\ell_1$ similarity with ground truth image. $\mathcal{L}_{sr}$ is the pose self-reconstruction loss introduced in \cite{DPTN}. $\lambda_{siadv}$, $\lambda_{style}$, $\lambda_{per}$, and $\lambda_{img}$ control their relative importance. 


\section{Experiments}
\noindent \textbf{Datasets:} The proposed approach is systematically experimented on three datasets with distinct characteristics. Since proposed gentle evolution synthesizing belongs to a relatively novel framework, in order to systematically explore its specific characteristics, we intentionally construct a new dataset, Turning-Round. This set is consisted of 28 persons, each of whom successively rotates in horizontal plane at fixed $15^\circ$ intervals. We randomly split all data in approximately 4:1 ratio, and got 23 people for training and 5 for evaluation. The large size Fashion dataset \cite{FashionData} characterized by various pose and clothes is adopted to comprehensively evaluate the overall pose synthesizing ability, where there are 500 training and 100 test people. In addition, in order to objectively assess generalization capacity, we further enroll the Tai-Chi dataset \cite{Tai-Chi}, which is constituted by 3000+ YouTube motion video clips (3049 training and 285 test). We sample at 3 FPS to collect the experimental images without any further interference. Hence, the collected poses for each individual are different with those of others, which means the evaluation experiments will be conducted on ''never trained'' poses. The same configurations of training and test samples are used for all compared approaches. 


\noindent \textbf{Evaluation Metrics:} Four commonly adopted image synthesizing quality indices are used for our experimental evaluation: Structural Similarity Index Measure (SSIM) \cite{SSIM}, Peak Signal to Noise Ratio (PSNR), Fr\'{e}chet
Inception Distance (FID) \cite{FID}, and Learned Perceptual Image Patch Similarity (LPIPS) \cite{LPIPS}.

\noindent \textbf{Implementation Details:} The experiments are conducted on one NVIDIA Tesla-A100 GPU. Our model is trained through the Adam optimizer \cite{Adam} with $\beta_1 = 0.5$ and $\beta_2 = 0.999$. In all experiments, the loss weights are uniformly set as:  $\lambda_{sadv}=1$, $\lambda_{ncons}=0.01$, $\lambda_{pose}=10$,  $\lambda_{siadv}=2$, $\lambda_{style}=500$, $\lambda_{per}=0.5$, and $\lambda_{img}=5$\footnote{In our experiments, the fluctuation of relative weights (less than 10\%) may incur at most [1.78\%(SSIM), 4.67\%(PSNR), 33.23\%(FID), 20.00\%(LPIPS)] performance degradation.}.   

\subsection{Verification Experiments}
\begin{figure}[t]
\centering
\includegraphics[width=\linewidth]{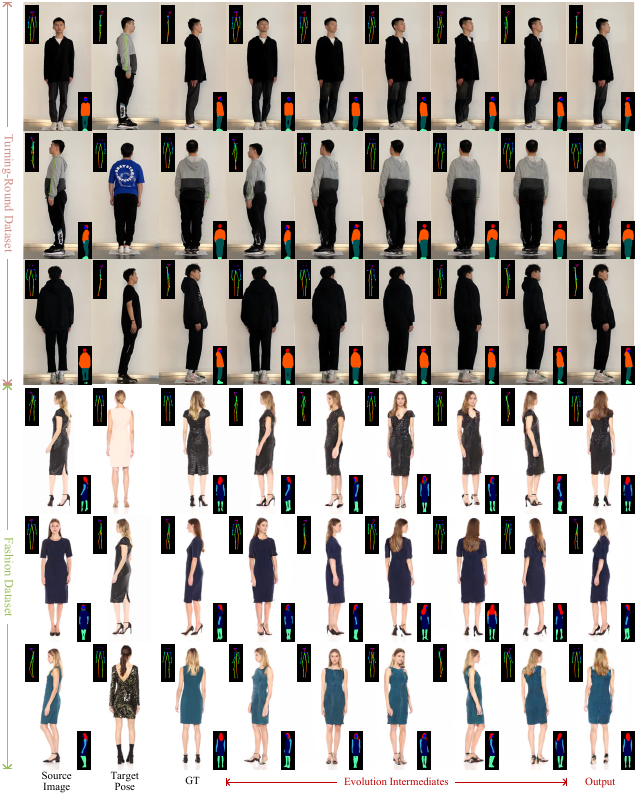}
\vspace{-5mm}
\caption{Synthesized poses and corresponding incremental evolution intermediates with skeleton and semantics on the Turning-Round and Fashion dataset.}
\label{fig:Turning-Round}
\vspace{-1mm}
\end{figure}

Figure \ref{fig:Turning-Round} demonstrates our generation performance (including synthesized skeleton and semantics) accompanied by corresponding evolution intermediates on both Turning-Round and Fashion dataset. It could be observed that proposed gentle incremental evolution pose generation framework effectively restrains the visual distortions and deformity easily incurred by direct dramatic variance modeling, and adequately maintains original visual texture. In addition, plenty of incremental intermediates could be acquired as qualified by-products. In big data era, obtain numbers of poses may facilitate a good many of related applications, such as 3D synthesizing, object recognition, and scene understanding.  
 

\begin{figure}[t]
\centering
\includegraphics[width=\linewidth]{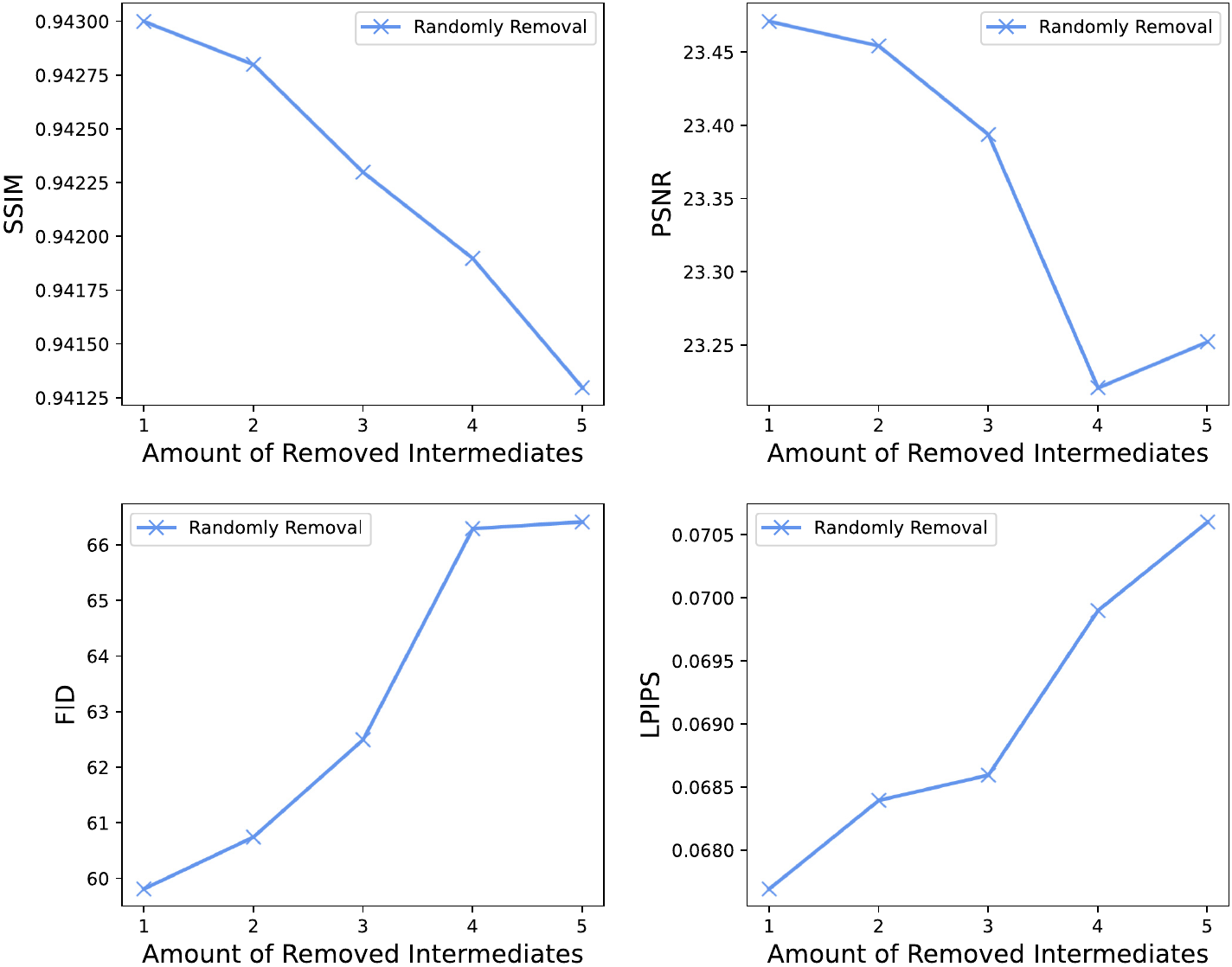}
\vspace{-5mm}
\caption{The pose synthesizing accuracy on the Turning-Round dataset when a few of evolution intermediate increments are randomly removed from the generation flow. }
\label{fig:removal}
\vspace{-4mm}
\end{figure}

The core underlying assumption of the whole proposed framework is that diminishing the extent of pose variation would relieve corresponding modeling difficulty, and that's why we update traditional rush one-to-one generation into an incremental gentle evolution course. Related verification experiments about this assumption are summarized in Table \ref{table:evo}, where the performance under various number of intermediate increments are explored. According to the table, increasing evolution increments could effectively improve generation accuracy, which directly supports our assumption. Similar phenomenon could also be found in Figure \ref{fig:removal}, where we randomly remove certain intermediates from the overall synthesizing flow. In the figures, there are evident general accuracy decrease accompanied by increments removal. 

\begin{table}[t]
    \footnotesize
    \begin{center}
    \resizebox{\linewidth}{!}{
        \begin{tabular}{lcccc}
          \toprule
            Increment Number   & SSIM$(\uparrow)$  & PSNR($\uparrow$)  & FID($\downarrow$)  & LPIPS($\downarrow$)  \\
         \midrule
    No Increments     & 0.936  & 22.615 & 66.159 & 0.075  \\
    One  Increment    & 0.939  & 22.989 & 66.228 & 0.073  \\
    Two  Increments   & \underline{0.940}  & \underline{23.223} & \underline{64.010} & \underline{0.072}  \\
    Five  Increments & \textbf{0.947}  & \textbf{23.730}  & \textbf{62.522} & \textbf{0.066}   \\
    \bottomrule
    \end{tabular}}
    \end{center}
\vspace{-3mm}
 \caption{The influence of evolutionary increment numbers to overall pose generation on the Turning-Round dataset.}
     \label{table:evo}
     \vspace{-4mm}
\end{table}

\vspace{-2mm}

\begin{table*}[t]
    \centering
     \renewcommand{\arraystretch}{0.85}
     \setlength{\tabcolsep}{1.1pt}
    \resizebox{\linewidth}{!}{
    \begin{tabular}{lccccccrcccrcrr}
    \toprule
    \multirow{2}{*}{Approaches} & \multicolumn{4}{c}{Turning-Round} & \multicolumn{4}{c}{Fashion} & \multicolumn{4}{c}{Tai-Chi} &\multicolumn{2}{c}{Overhead} \\\cmidrule(lr{0pt}){2-5}\cmidrule(lr{0pt}){6-9}\cmidrule(lr{0pt}){10-13}\cmidrule(lr{0pt}){14-15}
    \multicolumn{1}{c}{}
    &SSIM$(\uparrow)$ & PSNR$(\uparrow)$  & FID$(\downarrow)$  & LPIPS$(\downarrow)$  & SSIM$(\uparrow)$ & PSNR$(\uparrow)$ & FID$(\downarrow)$ & LPIPS$(\downarrow)$ & SSIM$(\uparrow)$ & PSNR$(\uparrow)$ & FID$(\downarrow)$ & LPIPS$(\downarrow)$ & \#Param$(\downarrow)$ & MACs$(\downarrow)$\\
    \midrule
    PATN \cite{PATN} (CVPR’19)               & 0.912 & 21.099 & 75.866 & 0.082 & 0.882  & 22.103 & 18.326 & 0.091  & 0.586  & 16.835 & 101.815 & 0.315  & 41.36 M & 187 G\\
    XingGAN \cite{XingGAN} (ECCV’20)         & 0.916 & 21.106 & 65.383 & 0.081 & 0.906  & 23.256 & 18.599 & 0.072  & 0.627  & 18.051 & 109.830 & 0.284  & 44.84 M & 265 G\\
    ADGAN \cite{AD-GAN} (CVPR’20)            & 0.929 & 22.267 & 63.822 & 0.070 & 0.899  & 22.879 & 23.657 & 0.079  & 0.285  & 13.577 & 126.955 & 0.556  & 48.79 M & 424 G\\
    PINet \cite{PINet} (CGF’20)              & 0.813 & 16.859 & 111.522& 0.147 & 0.916  & 23.998 & 15.678 & 0.065  & 0.645  & 19.113 & 96.411  & 0.270  & 20.41 M & 173 G\\
    PISE \cite{PISE} (CVPR’21)               & 0.948 & 23.863 & 57.541 & 0.058 & 0.900  & 22.571 & 20.967 & 0.098  & 0.401  & 14.186 & 105.627 & 0.516  & 64.01 M & 150 G\\
    SPGNet \cite{SPGNet} (CVPR’21)           & 0.949 & 23.949 & 56.853 & 0.056 & 0.921  & 24.898 & 16.757 & 0.051  & 0.611  & 18.396 & 97.840  & 0.265  & 87.79 M & 350 G\\
    CASD \cite{CASD} (ECCV’22)               & 0.924 & 21.870 & 64.610 & 0.071 & 0.893  & 22.709 & 20.950 & 0.075  & 0.301  & 14.272 & 129.788 & 0.536  & 58.51 M & 167 G\\
    DPTN \cite{DPTN} (CVPR’22)               & 0.933 & 22.146 & 51.719 & 0.064 & 0.896  & 22.806 & 14.070 & 0.071  & 0.576  & 17.024 & 99.758  & 0.307  & \textbf{9.79 M} & \textbf{61 G}\\
    BiGraphGAN \cite{BiGraphGAN} (IJCV’23)   & 0.921 & 21.490 & 90.746 & 0.081 & 0.907  & 23.298 & 13.524 & 0.709  & 0.668  & 19.331 & 85.746  & 0.234  & 42.10 M & 191 G\\
    MAGPT \cite{MSAGPT} (PR'23)              & 0.927 & 21.621 & 57.860 & 0.069 & 0.898  & 22.605 & 17.951 & 0.073  & 0.643  & 18.641 & 106.875 & 0.299  & 90.17 M & \underline{130 G}\\
    PIDM \cite{PIDM} (CVPR’23)               & 0.703 & 14.775 & 59.275 & 0.190 & 0.890 & 22.319  & \underline{12.899} & 0.067 & 0.599 & 17.515 & 84.391 & 0.237  & 131.14 M & 27,940 G\\
    \midrule
    Ours: Evolution View-S & 0.951 & 24.179 & 53.271 & 0.055 & 0.927  & 24.919 & 14.394 & \underline{0.050}  & 0.681  & 19.485 & 76.436 & 0.235  & \underline{11.08 M} & 131-457 G\\
    Ours: Evolution View-B & \underline{0.952} & \underline{24.180} & \underline{53.328} & \underline{0.054} & \underline{0.928}  & \underline{24.959} & 14.002 & \underline{0.050}  & \underline{0.684}  & \underline{19.516} & \underline{76.838} & \underline{0.228}  & 13.19 M & 132-461 G\\
    Ours: Evolution View-L & \textbf{0.953} & \textbf{24.265} & \textbf{51.384} & \textbf{0.053} & \textbf{0.930}  & \textbf{25.332} & \textbf{12.720} & \textbf{0.048}  & \textbf{0.689}  & \textbf{19.705} & \textbf{75.591} & \textbf{0.227} & 15.30 M & 133-464 G\\
    \bottomrule
    \end{tabular}
    }
    \vspace{-1.5mm}
        \caption{Quantitative comparison of pose synthesis quality on the Turning-Round, Fashion, and Tai-Chi dataset.}
            \label{table:cmp}
               \vspace{-4mm}
\end{table*}

\subsection{Comparison with State of the Art Methods}
The detailed quantitative comparison with other SOTA approaches are summarized in Table \ref{table:cmp}, where three our structures are specifically explored: -S, -B, and -L, which respectively correspond to two, four and six stacked Triple-Path Knowledge Fusion and Source Feature Extraction blocks. It could be observed from the table that proposed approaches achieve competitive performance compared with other SOTA approaches. Furthermore, even the smallest size -S could outperform most of other SOTA methods. In addition, our model size is better than most of other approaches (only higher than DPTN), as to parameter amounts. But since we inherently integrate an evolution course, our MACs (Multiply-ACcumulate operations) is not quite well: from 2 to 7 increments configurations, the corresponding MACs ranges from 130G+ to 450G+, an average moderate and lower level. But we could simultaneously synthesize by-products as compensation to the incurred computing costs.    

\begin{figure}[t]
\centering
\includegraphics[width=\linewidth]{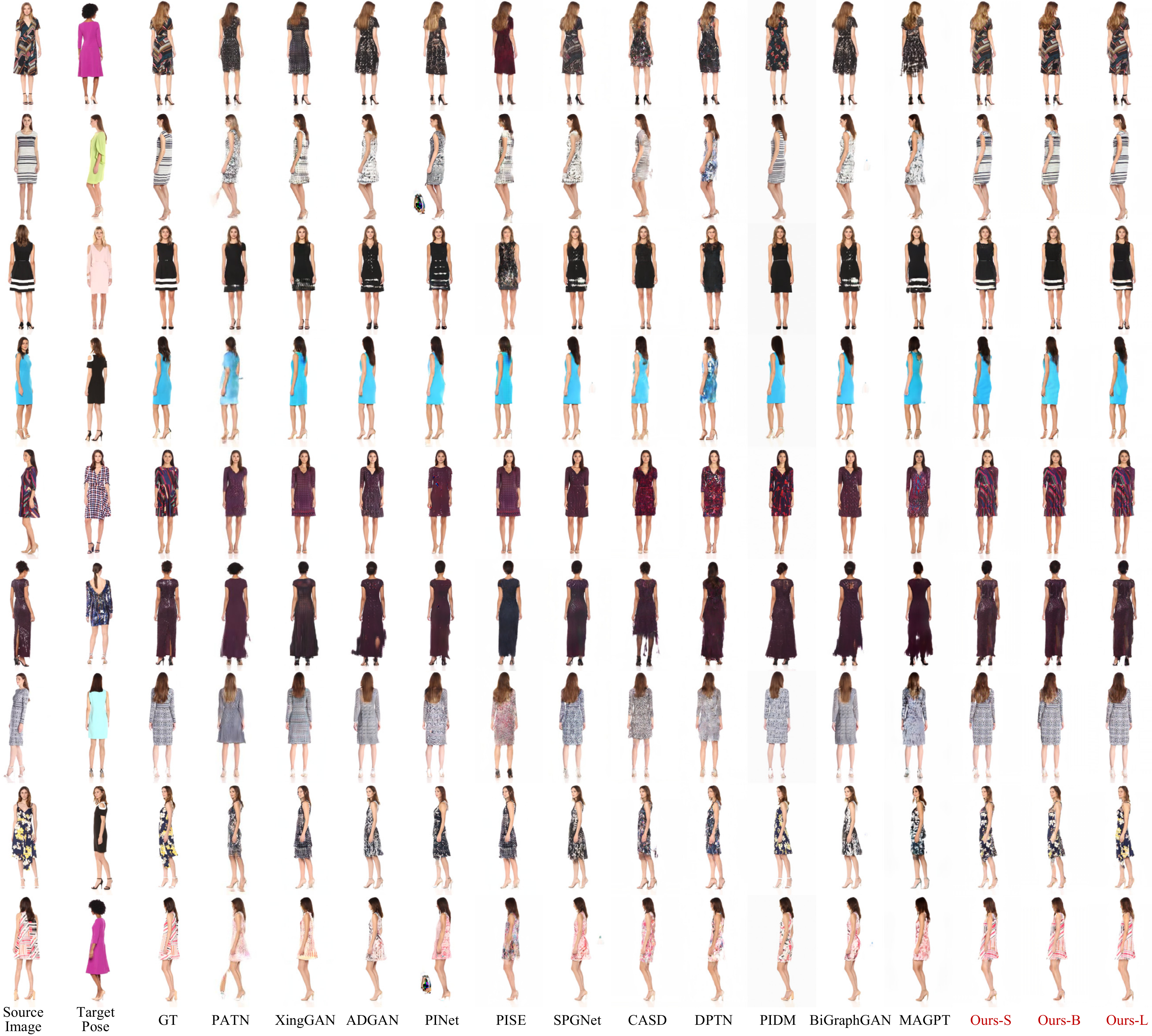}
\vspace{-6mm}
\caption{Qualitative comparison of pose synthesis on the Fashion dataset. Please zoom in for better view.}
\label{fig:fashion}
\vspace{-1mm}
\end{figure}
\begin{figure}[!h]
\centering
\includegraphics[width=\linewidth]{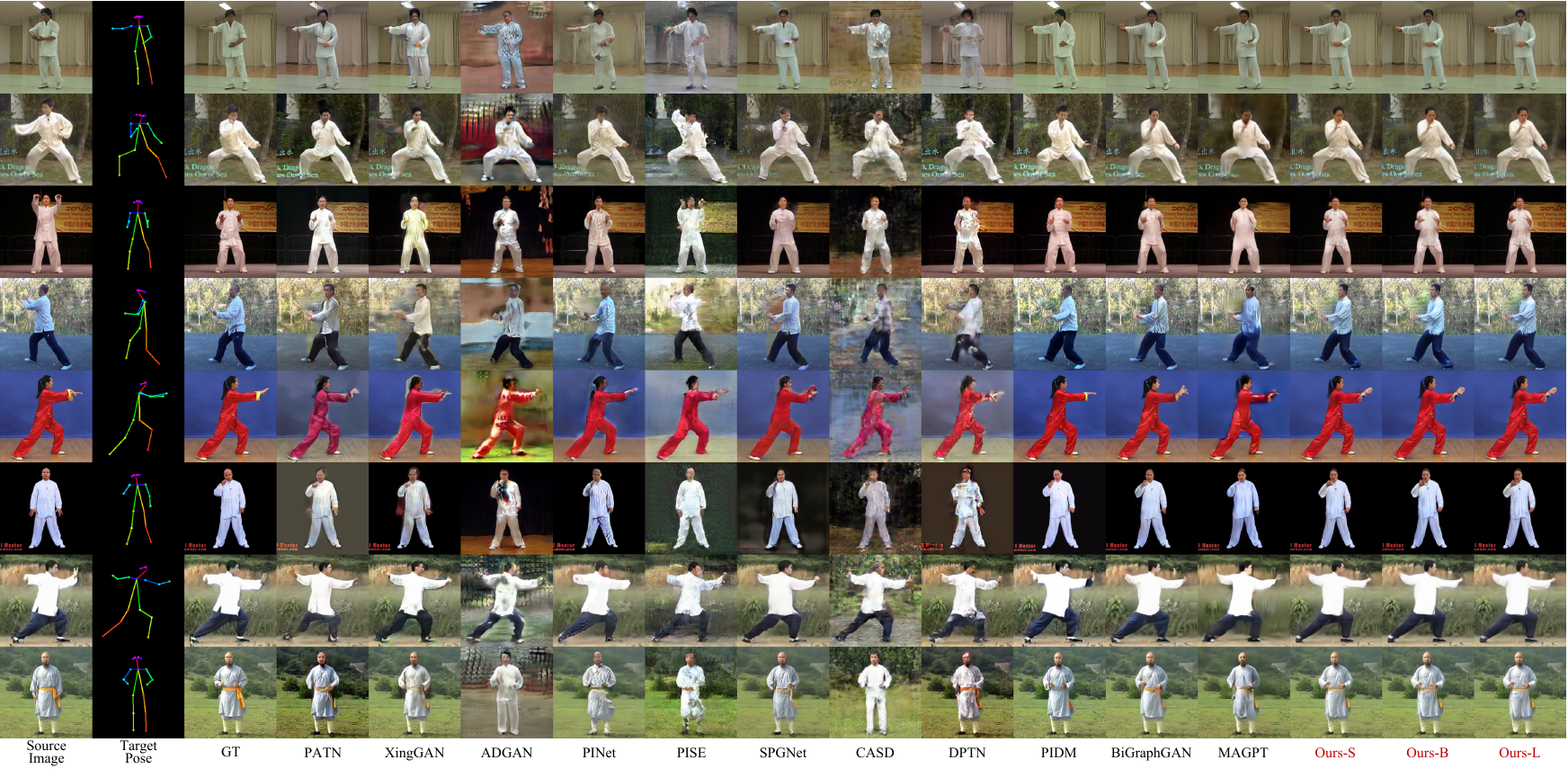}
\vspace{-6mm}
\caption{Qualitative comparison of pose synthesis towards skeleton pose targets on the Tai-Chi dataset. Please zoom in for better view.}
\label{fig:taichi}
\vspace{-4mm}
\end{figure}
\begin{figure}[!h]
\centering
\includegraphics[width=\linewidth]{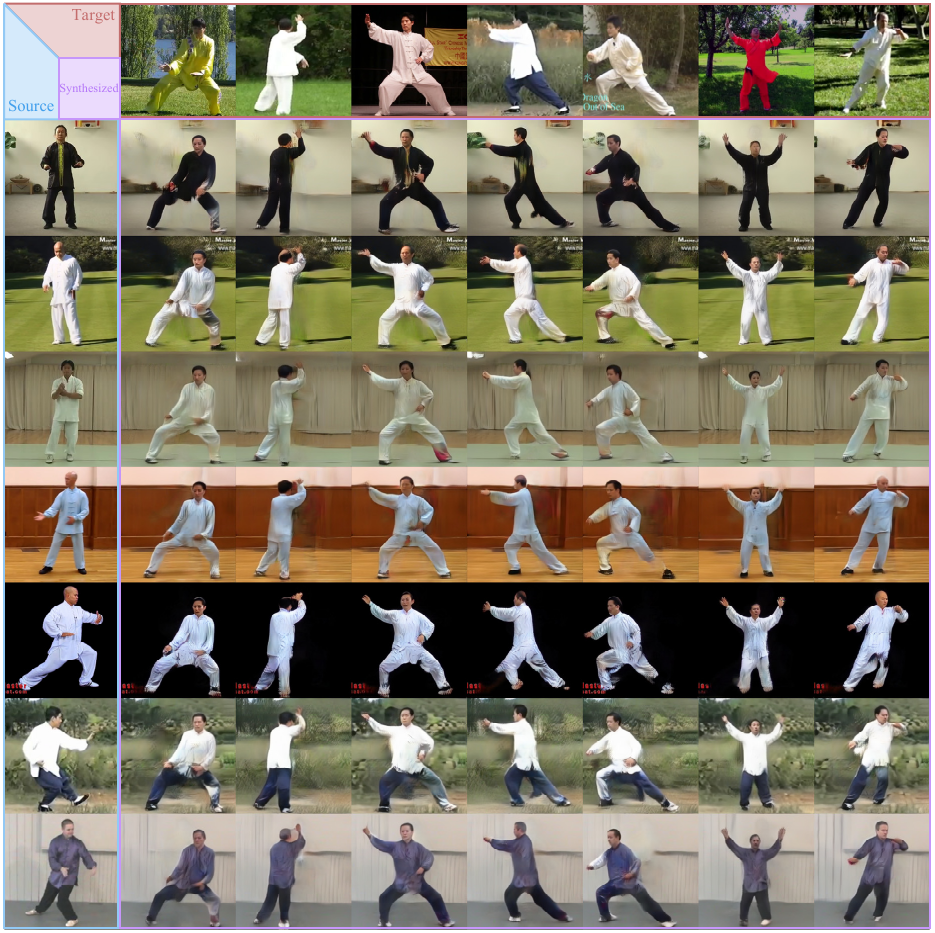}
\vspace{-6mm}
\caption{Qualitative synthesizing performance towards target poses of other subjects on the Tai-Chi dataset. }
\label{fig:taichiother}
\end{figure}

The qualitative comparison is illustrated in Figure \ref{fig:fashion} and \ref{fig:taichi}. It should be mentioned that since the poses of each individual in the Tai-Chi dataset are different, we had to conduct synthesis towards the skeleton targets of their own. Otherwise, there will be no ground truth reference for comparison. As supplement, we also experimented the generation performance directly towards pose targets (without ground truth) in Figure \ref{fig:taichiother}. It shows in these figures that proposed approach could adequately realize dramatic variance pose transformation. The details, especially visual textures, could be properly maintained without incurring content distortion. Furthermore, the intrinsic individual pose distinctions in the Tai-Chi dataset imply fine generalization capacity could be achieved: when dealing with poses never shown in training set, proposed approach could still generate qualified target poses (Figure \ref{fig:taichiother}).   

\subsection{Human Perceptual Study}
Since human feelings are still the most critical metrics to evaluate an image synthesizing strategy, we also conduct human perceptual experiments. Specifically, 25 ground truth and 25 synthesized images are randomly picked out from the Fashion dataset for feeling evaluation by 25 volunteers. Similar with \cite{CASD}, we enroll both R2G and G2R as measurements. R2G: the percentage of the real images recognized as generated; G2R: the percentage of the generated images recognized as real. The detailed performance is shown in Table \ref{table:Human}. Here, it could be observed that proposed approach could achieve better user feelings compared with other approaches.  

\begin{table}[t]
    \centering
    \footnotesize
               \resizebox{0.9\linewidth}{!}{
                   \begin{tabular}{lcccc} 
                      \toprule
                     \multirow{2}{*}{Metrics}                  &  PISE             &CASD                & BiGraphGAN        &\multirow{2}{*}{Ours}        \\
                                                                                  &(CVPR’21)     &(ECCV’22)        & (IJCV’23)                                                                \\\midrule
                      R2G$(\uparrow)$   & 16.8\%                         &27.1\%                                & 38.8\%                                     &\textbf{42.4\%}  \\ 
                      G2R$(\uparrow)$   & 11.2\%                         &12.0\%                                & 13.8\%                                     & \textbf{20.8\%}\\\bottomrule               
                  \end{tabular}
                  }
   \vspace{-1.5mm}
                   \caption{Human perceptual study on the Fashion dataset.}
                      \label{table:Human}
   \vspace{-4mm}
\end{table}

\subsection{Ablation Study}
Table \ref{table:ablation} detailed demonstrates the value of core components/operations to the proposed framework. It could be found from the table that both TPKF and IEC are important to the overall synthesis performance. Their absence may incur at most [2.48\%(SSIM), 6.06\%(PSNR), 5.90\%(FID), 21.88\%(LPIPS)] performance degradation. The removal of the multi-scale convolution within Incremental Evolution blocks, and the extra AdaIN mechanism of Triple-Path Knowledge Fusion blocks can cause moderate accuracy reduction, namely at most [1.08\%(SSIM), 3.31\%(PSNR), 1.77\%(FID), 13.79\%(LPIPS)]. The value of face refinement component to the overall quantitative accuracy is slightly, namely [0.11\%(SSIM), 0.05\%(PSNR), 0.28\%(FID), 1.96\%(LPIPS)]. On the other hand, according to the lower part of the table, solo stacking more Incremental Evolution blocks could not improve synthesizing performance, and even certain adverse effects could be observed. Here, we believe that's because too deeper convolution fusion may lead to critical details lost. 

\begin{figure}[!h]
\centering
\includegraphics[width=\linewidth]{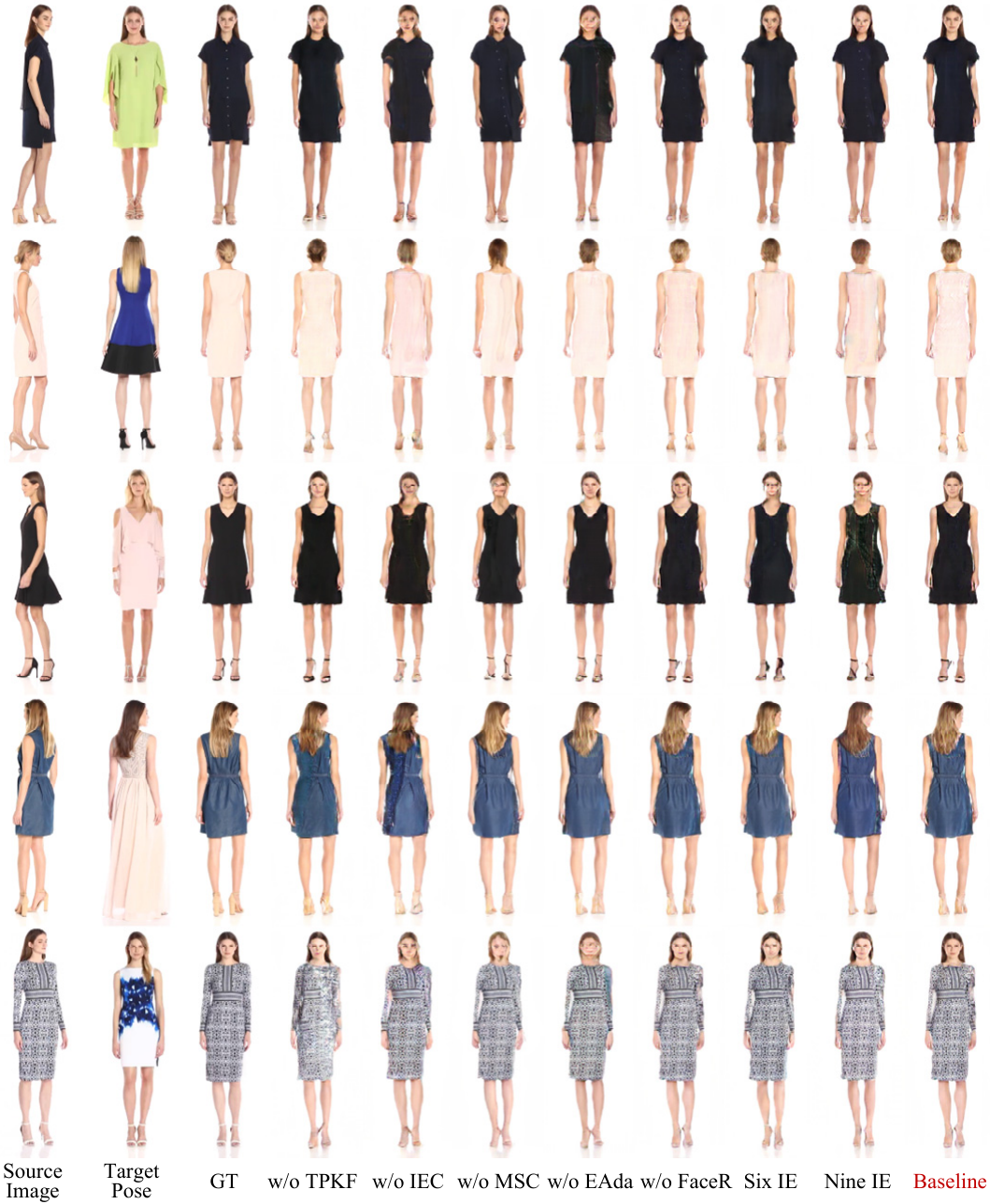}
\vspace{-6mm}
\caption{Qualitative ablation study on the Fashion dataset.}
\label{fig:Ablation}
\vspace{-3mm}
\end{figure}

Corresponding qualitative performance is demonstrated in Figure \ref{fig:Ablation}. Here, it could be found that without these key components/operations, the synthesized poses may be degraded to some extent.  


\begin{table}
    \begin{center}
    \resizebox{\linewidth}{!}{
    \begin{tabular}{lcccc}
    \toprule
        \multicolumn{1}{l}{Model} 
    & SSIM$(\uparrow)$  & PSNR$(\uparrow)$  & FID$(\downarrow)$  & LPIPS$(\downarrow)$  \\
    \midrule
    w/o Triple-Path Knowledge Fusion (TPKF) & 0.904  & 23.410 & 15.296 & 0.059   \\
    w/o Incremental Evolution Constraints (IEC)   & 0.909  & 23.689 & 14.676 & 0.064  \\
    w/o Multi-Scale Convolution (MSC) & 0.917  & 24.093  & 14.645 & 0.058   \\
    w/o Extra AdaIN (EAda) & 0.922  & 24.599  & \textbf{14.135} & 0.053 \\
    w/o Face Refinement (FaceR) & \underline{0.926}  & \underline{24.906} & 14.434 & \underline{0.051} \\
    \midrule
    Six Stacked IE Blocks (Six IE) & 0.921  & 24.454 & 15.197 & 0.055  \\
    Nine Stacked IE Blocks (Nine IE) & 0.914  & 23.857 & 15.269 & 0.062  \\
    \midrule
    Our Baseline &  \textbf{0.927}  &  \textbf{24.919} &  \underline{14.394} &  \textbf{0.050} \\
    \bottomrule
    \end{tabular}
           
    }
    \end{center}
    \vspace{-3.5mm}
        \caption{Quantitative ablation study on the Fashion dataset. }
            \label{table:ablation}
   \vspace{-1mm}
\end{table}

\section{Conclusion}
In this paper, to accurately conduct robust human pose synthesizing, unlike traditional one-to-one rush transformation, we designed a slight pose transformation modeling unit centered gentle incremental evolution framework, which is a novel way to handle the theoretically difficult mission of modeling huge non-linear visual content discrepancy. 

In order to rigorously control the evolution course to achieve high-quality ultimate output, both global and incremental evolution constraints are imposed, which strictly supervise and guide the overall operation flow. Furthermore, we propose a triple-path knowledge fusion mechanism to make full use of all available valuable knowledge for favorable synthesizing performance. In addition, besides the prescriptive target pose, a series of valuable by-products, namely the various intermediate poses, could also be acquired. This may be an extra compensation to our relatively ordinary computing overhead. Both quantitative and qualitative experiments have demonstrated that evident accuracy advantages could be achieved compared with other SOTA approaches.

\section{Acknowledgments}
This work was supported in part by the National Natural Science Foundation of China under Grant 61771340, 62072335, and 62176182.



\bibliography{aaai24}

\end{document}